\documentclass{ws-ijwmip}
\usepackage{graphicx}
\usepackage{subfigure}
\usepackage{amsmath,amssymb}
\usepackage{multirow}
\usepackage{color}
\usepackage{xcolor}
\usepackage{algorithm}
\usepackage{algorithmic}
\usepackage{float}
\usepackage{makecell}
\usepackage[super]{cite}
\usepackage{hyperref}
\usepackage{booktabs}
\usepackage{CJKutf8}
\usepackage[misc,geometry]{ifsym}

\begin{document}
\begin{CJK}{UTF8}{gbsn}

\markboth{Authors' Names}{Paper's Title}

\catchline{}{}{}{}{}

\title{Semantic Guided Level-Category Hybrid Prediction Network for Hierarchical Image Classification}

\author{Peng Wang}
\address{College of Computer Science, Zhejiang University, Hangzhou, Zhejiang 310007, China\\
pengwang18@zju.edu.cn}

\author{Jingzhou Chen}
\address{Ant Group, Hangzhou, Zhejiang 310007, China\\
chenjingzhoucs@zju.edu.cn}

\author{Yuntao Qian\footnote{Corresponding author}}
\address{College of Computer Science, Zhejiang University, Hangzhou, Zhejiang 310007, China\\
ytqian@zju.edu.cn}

\maketitle

\begin{history}
\received{(Day Month Year)}
\revised{(Day Month Year)}
\accepted{(Day Month Year)}
\published{(Day Month Year)}
\end{history}

\begin{abstract}
    Hierarchical classification (HC) assigns each object with multiple labels organized into a hierarchical structure. The existing deep learning based HC methods usually predict an instance starting from the root node until a leaf node is reached. However, in the real world, images interfered by noise, occlusion, blur, or low resolution may not provide sufficient information for the classification at subordinate levels. To address this issue, we propose a novel Semantic Guided level-category Hybrid Prediction Network (SGHPN) that can jointly perform the level and category prediction in an end-to-end manner. SGHPN comprises two modules: a visual transformer that extracts feature vectors from the input images, and a semantic guided cross-attention module that uses categories word embeddings as queries to guide learning category-specific representations. In order to evaluate the proposed method, we construct two new datasets in which images are at a broad range of quality and thus are labeled to different levels (depths) in the hierarchy according to their individual quality. Experimental results demonstrate the effectiveness of our proposed HC method.
\end{abstract}



\section{Introduction}  

\begin{figure}[t]
    \centering
    \subfigure[Birds with/without occlusion.]{   
    \includegraphics[width=2.43in]{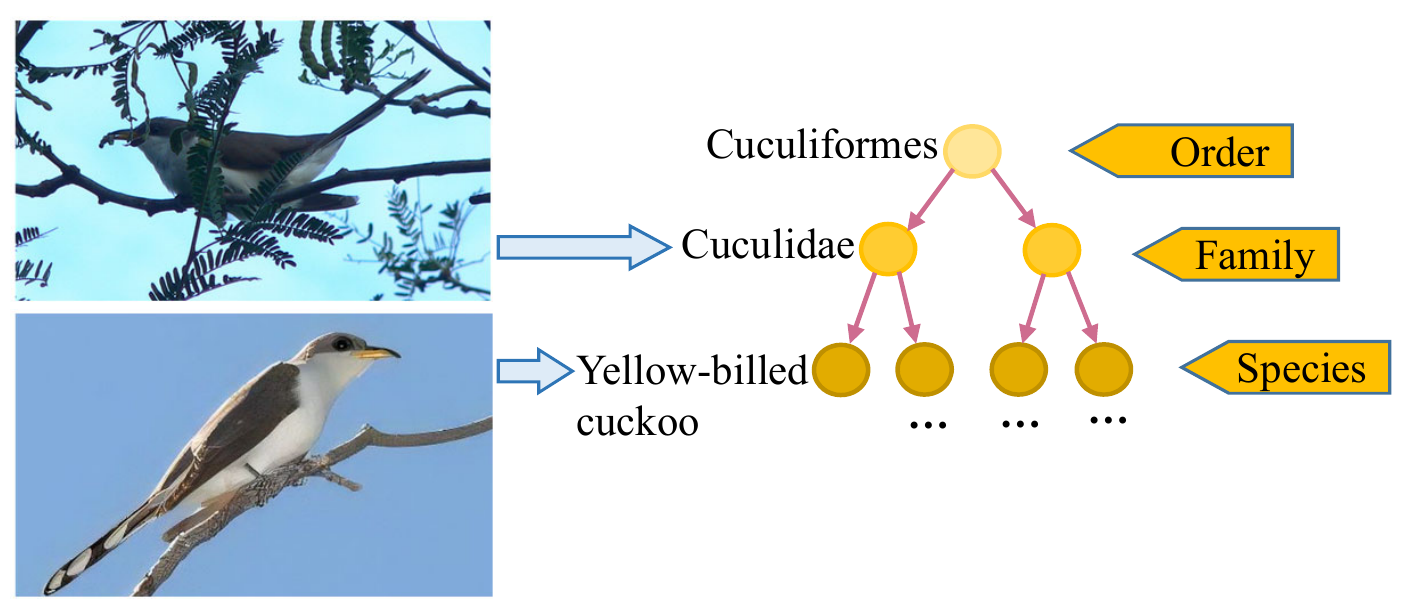}}
    \subfigure[Ship images of low/high resolution.]{
    \includegraphics[width=2.43in]{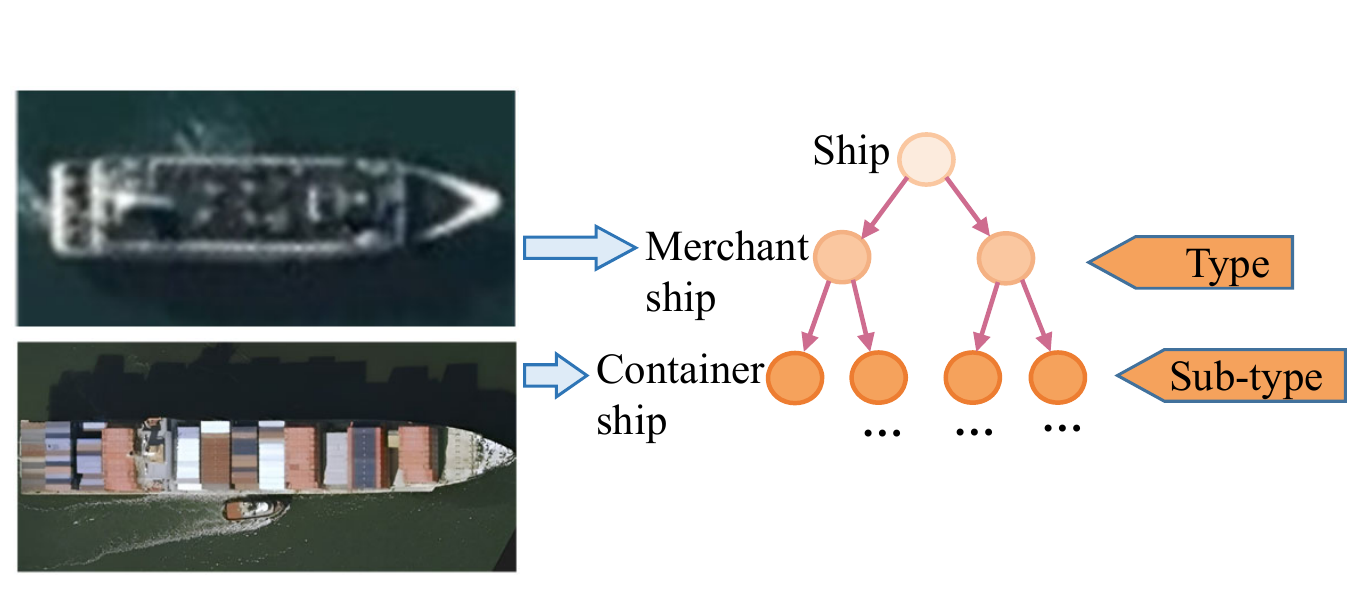}}\\
    \caption{Objects can be perceived at varied levels in the label hierarchy due to the variation of image quality.}
    \label{motivation_fig}
\end{figure}

Recently, hierarchical classification (HC) has attracted increasing attention in visual recognition~\cite{Gopal2015HierarchicalBI, luo2015image, Hoyoux2016CanCV, Zhao2017HierarchicalFS, giunchiglia2020coherent}. However, most of the existing deep learning based HC methods~\cite{fan2017hd, zhao2018embedding, chen2022label, chen2021hierarchical, wehrmann2018hierarchical, cerri2016reduction, chang2021your} predict an image from the root to the leaf node, without considering whether sufficient information is provided for classification at lower levels (\textit{i.e.}, fine-grained levels). This poses a great risk of misclassification because, in the real world, images are often impaired by noise, occlusion, blur, or low resolution, as shown in Fig.~\ref{motivation_fig}, which makes it infeasible for prediction at lower levels. To reflect this practical issue, we assume that each image should belong to the appropriate level in the class hierarchy according to its image quality. Accordingly, we propose a novel Semantic Guided level-category Hybrid Prediction Network (SGHPN) for hierarchical classification that jointly performs the level and category prediction. It consists of two modules: 1) a visual transformer~\cite{dosovitskiy2020image} that maps an input image to a sequence of token embeddings; 2) a semantic guided cross-attention module (SGCA) that uses the token embeddings as keys and values, and uses the category semantics from label hierarchy as queries to mine the semantic-aware token embeddings and extract the category-specific features via cross-attention learning. 

In order to simulate the HC situation mentioned above and evaluate the proposed method, datasets in which images are of varied quality and thus are labeled to different levels are necessary. The existing image datasets used in HC usually have high-quality samples labeled from the top to the leaf level. Thus, these datasets are unsuitable for the proposed scenario and drive us to construct new HC image datasets. Since collecting such datasets from scratch can be prohibitively expensive, we propose a relatively cheap and efficient way that leverages images from existing datasets. To render the images varied in visual quality, we distort these images using four different image distortions, \textit{i.e.}, white Gaussian noise, motion blur, downsample, and cutout. Then we annotate these distorted images to different levels (depths) of the label hierarchy according to their individual image quality.

We summarize our main contributions as follows:
\begin{itemize}
    \item An HC method is proposed that predicts both the classes and the level of a sample in an end-to-end manner. To effectively utilize the semantics of categories in the label hierarchy, we propose semantic guided cross-attention module.
    \item Two novel datasets are constructed for hierarchical classification in which images are at a broad range of quality and thus are labeled to different levels in the hierarchy according to their varied quality.
\end{itemize}

\section{Datasets}

In this section, we aim to construct visual datasets for hierarchical classification where the images are varied in quality and thus are labeled to different levels in the category hierarchy. Collecting such datasets from scratch can be prohibitively expensive. A relatively cheap way is to leverage the existing datasets. Considering that the fine-grained visual classification datasets have class hierarchy and their images are of high quality, they are ideal sources for us to make new benchmarks. To render the images varied in quality, we propose to distort the source images using four different image distortions. 

Two commonly used fine-grained visual classification datasets are adopted as our source datasets: the CUB-200-2011 dataset~\cite{wah2011caltech} and the HRSC dataset~\cite{liu2017high}. The CUB-200-2011 dataset collects 11788 high-quality images of birds organized into 13 orders, 38 families, and 200 species. There are roughly 60 images for each species. This dataset provides a bounding box for each image and contains part annotations of 15 bird parts, \textit{e.g.}, beak, crown, tail, \textit{etc.}

The HRSC dataset is widely used in remote sensing ship classification, whose image size ranges from 300$\times$300 to 1500$\times$900. It contains 617 training images and 438 test images organized into 3 coarse-grained classes and 21 fine-grained classes. As the HRSC dataset provides bounding box annotations and the ship instance often only occupies a small part of the whole image, we crop every ship instance in images to avoid interference from the background. Since the new datasets are derived from CUB-200-2011 dataset and HRSC dataset, we denote them as CUB-200-2023 and HRSC-2023, respectively.

\subsection{Image Distortion Types}

\begin{figure*}[h]
    \centering
    \subfigure[Source image]{
    \includegraphics[width=1.11in]{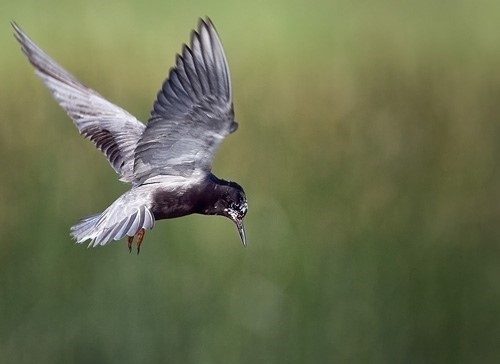}}
    \subfigure[Distorted by white noise]{
    \includegraphics[width=1.11in]{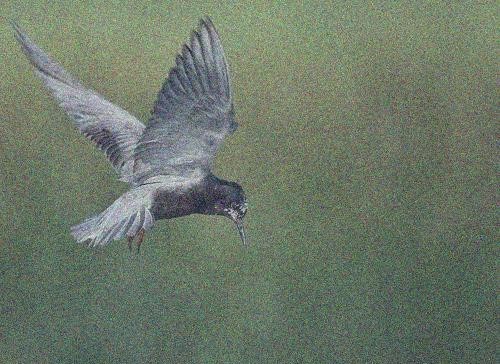}}
    \subfigure[Source image]{
    \includegraphics[width=1.11in]{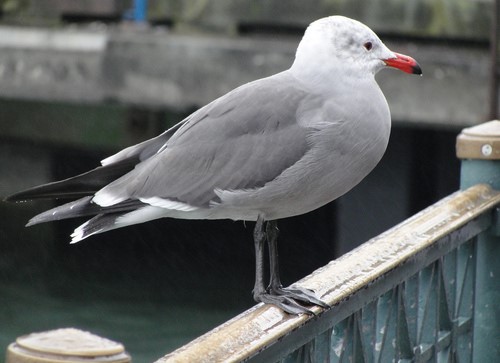}}
    \subfigure[Distorted by motion blur]{
    \includegraphics[width=1.11in]{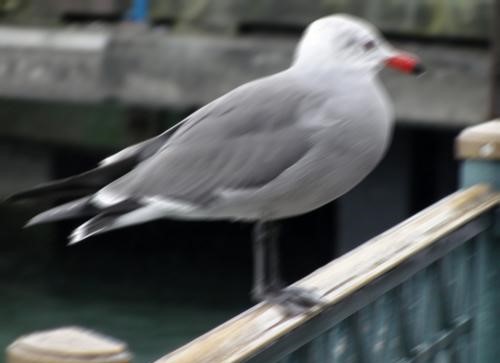}}
    \subfigure[Source image]{
    \includegraphics[width=1.11in]{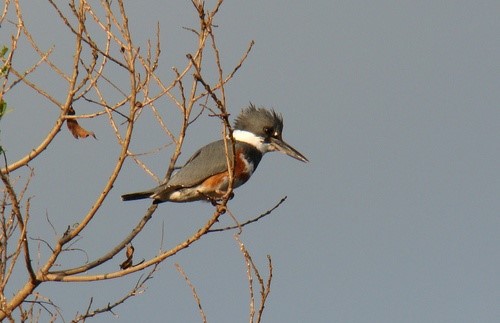}}
    \subfigure[Distorted by downsampling]{
    \includegraphics[width=1.11in]{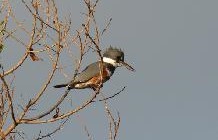}}
    \subfigure[Source image]{
    \includegraphics[width=1.11in]{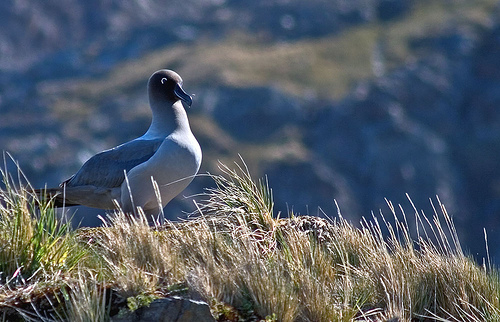}}
    \subfigure[Distorted by cutout]{
    \includegraphics[width=1.11in]{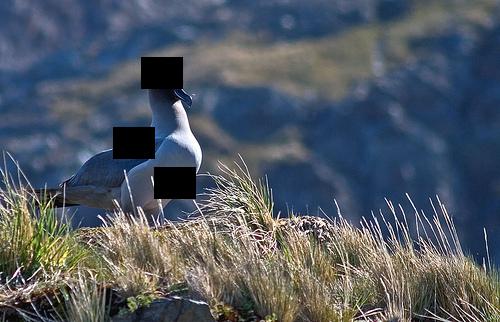}}
    \subfigure[Source image]{
    \includegraphics[width=1.11in]{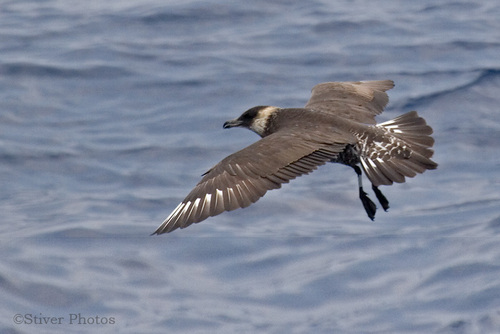}}
    \subfigure[Impaired by multiple distortions]{
    \includegraphics[width=1.11in]{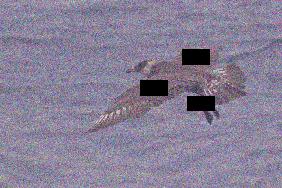}}
    \subfigure[Source image]{
    \includegraphics[width=1.11in]{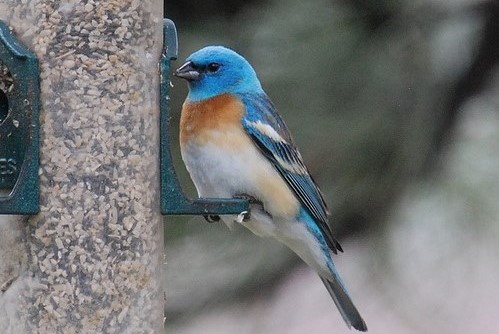}}
    \subfigure[Impaired by multiple distortions]{
    \includegraphics[width=1.12in]{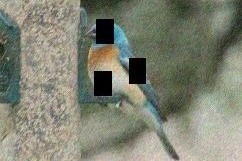}}\\
    \caption{The illustrative examples of source images from CUB-200-2011 dataset and distorted images from our CUB-200-2023 dataset. The category labels and level labels of the distorted images and the corresponding distortions are listed in Table~\ref{labels_table}.}
    \label{distort_fig}
\end{figure*}

Four distortions are chosen to degrade the source images:

\begin{itemize}
  \item \emph{White noise}. The white Gaussian noise of standard deviation $\sigma$ is added to the R, G, and B channels of the images. The same $\sigma$ is used for the three color channels, with a value between 0.1 and 0.2. 
  Fig.~\ref{distort_fig}(b) exemplifies the white noise.

  \item \emph{Motion blur}. The image's R, G, and B channels are filtered using a directional Gaussian kernel. The direction is randomly chosen between 0$^{\circ}$ and 180$^{\circ}$. The kernel size $\eta$ is used to control the blur strength, which is between 5 and 14. The three color channels of the image were blurred using the same kernel. Fig.~\ref{distort_fig}(d) is an illustrative example of motion blur image. 

  \item \emph{Downsampling}. We downsample the source images to simulate the low-resolution images. The downsample rate $\lambda$, defined as the ratio of the output image's area to the source image's area, is between 0.4 and 0.7. Fig.~\ref{distort_fig}(f) is an illustrative example of a downsampled image.

  \item \emph{Cutout}~\cite{Devries2017ImprovedRO}. Occlusions are commonly encountered in images. 
  Such occlusions pose great challenges in object classification. To simulate occlusions, we propose to randomly mask out a part of the object in the image. Specifically, for the CUB-200-2023 dataset, thanks to the off-the-shelf bounding box and parts locations, we randomly choose three bird parts and apply fixed-size zero masks to them. The shape of the mask patch is similar to the bounding box. We use the ratio of the mask patch area to the bounding box area to control the cutout strength, denoted as $\delta$. Fig.~\ref{distort_fig}(h) presents an example image after cutout. For the HRSC-2023 dataset, since the outline and the center part of the ship usually play an important role in ship classification, we randomly choose four locations among the center, left, right, top, and bottom of the ship to apply the zero mask.
\end{itemize}

Among these four distortions, one or multiple distortions are randomly applied to each source image (see Fig.~\ref{distort_fig}), and the strength of each distortion is varied to generate images of a broad range of quality.

\begin{table}[t]\footnotesize
    \caption{The distortions and labels corresponding to the distorted images in Fig.~\ref{distort_fig}.}
    \label{labels_table}
    \centering
    {\begin{tabular}{@{}ccccccc@{}} 
        \toprule
        Distorted images          & $\sigma$ & $\eta$    & $\lambda$ & $\delta$& Category label & Level label \\ 
        \hline
        Fig.~\ref{distort_fig}(b) & 0.14     & -         & -        & -        & [3, 4, 141]     & species \\
        Fig.~\ref{distort_fig}(d) & -        & 10        & -        & -        & [3, 4, 60]      & species \\ 
        Fig.~\ref{distort_fig}(f) & -        & -         & 0.44     & -        & [4, 6, None]    & family \\
        Fig.~\ref{distort_fig}(h) & -        & -         & -        & 0.04     & [11, 34, None]  & family \\ 
        Fig.~\ref{distort_fig}(j) & 0.19     & -         & 0.56     & 0.03     & [3, None, None] & order \\
        Fig.~\ref{distort_fig}(l) & 0.15     & 5         & 0.49     & 0.03     & [7, None, None] & order \\
        \botrule
    \end{tabular}}
\end{table}

\subsection{Data Annotation}

In this subsection, we describe how to annotate the samples in the introduced datasets. Since the number of samples and categories is large, it can be prohibitively expensive and time-consuming for human beings to annotate the datapoints. A relatively cheap and efficient way is to leverage the state-of-the-art HC model to annotate the datapoints because the deep learning based models now show remarkable ability in recognizing objects in images. As the model presented in the work of Chang et al.\cite{chang2021your} is claimed to establish state-of-the-art performances in hierarchical classification, we adopt it as our annotator. Notably, other state-of-the-art models are also suitable for this task.

Two kinds of labels are necessary for the new datasets: (1) the category label that denotes the image's class, and (2) the level label that indicates which level the image belongs to. Since the source datasets have provided the category labels, our annotator model only needs to annotate level labels. In practice, we train the annotator model with all \emph{source} images from CUB-200-2011 dataset or HRSC dataset. Then this trained model is frozen and is used to annotate the \emph{distorted} images, \textit{i.e.}, tag each image with a level label. Empirically, the annotator model is more likely to make misclassification when it goes more deeply in the class hierarchy. Therefore, we take the finest-grained level that the annotator model can classify correctly for a distorted image as this image's ground truth level. For example, for a distorted image in CUB-200-2023 dataset, if the annotator model classifies it correctly at order and family level, but wrongly at species level, then the ground truth level of this image is family.

However, there are some cases where the annotator model's predictions disobey hierarchical constraints. For a label tree hierarchy as adopted in our datasets, the hierarchical constraints require that a sample belonging to a given class must also belong to all the ancestor classes in the hierarchy~\cite{giunchiglia2020coherent}. In other words, if a sample is classified correctly in subordinate levels, it must be classified correctly in superordinate levels. For CUB-200-2023 dataset and HRSC-2023 dataset, there are respectively three cases and two cases that adhere to the hierarchical constraints, which are presented in Table~\ref{legal_set_table}. These cases are simplified as a legal set $s$, which is shown in the last row of Table~\ref{legal_set_table}. Algorithm 1 provides the procedures for the construction and annotation of new datasets. Formally, we denote the distortions as a function $G(t, \sigma, \eta, \lambda, \delta)$, where the $t$ stands for the randomly chosen distortion types and $\sigma, \eta, \lambda, \delta$ have been defined in the last subsection. Input the source image $I_s$ into $G$, we obtain the distorted image $I_d$. If $I_d$'s prediction result respects the hierarchical constraints, we add $I_d$ into the new dataset. Otherwise, we repeat these steps until $I_d$'s prediction satisfies the hierarchical constraints.

\begin{table}[t]\footnotesize
    \caption{The cases that adhere to the hierarchical constraints in CUB-200-2023 and HRSC-2023 datasets. These cases are denoted as legal cases, which are simplified as a legal set $s$ for convenience. ``1'' means correct classification, and ``0'' means misclassification.}
    \label{legal_set_table}
    \centering
    {\begin{tabular}{r|ccc|cc} 
        \bottomrule
        &\multicolumn{3}{c|}{CUB-200-2023} & \multicolumn{2}{c}{HRSC-2023}  \\
        \cline{2-6} 
                                        & Order    & Family    & Species     & Type     & Sub-type    \\
                                        \hline
        \multirow{3}{*}{Legal cases}  & 1        & 1         & 1           & 1        & 1           \\ 
                                        \cline{2-6}
                                        & 1        & 1         & 0           & 1        & 0           \\ 
                                        \cline{2-6}
                                        & 1        & 0         & 0           & -        & -        \\ 
        \hline
        Legal set &\multicolumn{3}{c|}{$s=([1, 1, 1], [1, 1, 0], [1, 0, 0])$}  &\multicolumn{2}{c}{$s=([1, 1], [1, 0])$}     \\
        \toprule
    \end{tabular}}
\end{table}

\begin{figure}[t]
    \label{algor}
    \renewcommand{\algorithmicrequire}{\textbf{Input:}}
    \renewcommand{\algorithmicensure}{\textbf{Output:}}
    \begin{algorithm}[H]
        \algsetup{linenosize=\small} \small
        \caption{\small{Pseudocode of making datasets}}
        \begin{algorithmic}[1]
        \REQUIRE source dataset $\mathcal{D}_s$, the trained annotator model $f$, the distortion function $G$, legal set $s$. 
        \ENSURE new dataset $\mathcal{D}_n$.
        \FOR{$I_s \in \mathcal{D}_s$}
        \STATE randomly generate parameters $t, \sigma, \eta, \lambda, \delta$ for $G$;
        \STATE $I_d \leftarrow G(I_s)$;
        \STATE $p \leftarrow f(I_d)$;  \textcolor{gray}{\# For CUB-200-2023 dataset, prediction $p$ is in the form of $[p_1, p_2, p_3]$, where $p_i$ is either 1 or 0. For HRSC-2023 dataset, $p$ is in the form of $[p_1, p_2]$.}
        \IF{$p \in s$}
        \STATE add $I_d$ into $\mathcal{D}_n$;
        \ELSE 
        \STATE repeat step 2$\sim$8;
        \ENDIF
        \ENDFOR 
        \end{algorithmic}
    \end{algorithm}
\end{figure}

\begin{table}[t]\footnotesize
    \caption{Statistics of CUB-200-2023 and HRSC-2023 datasets.}
    \label{statistic_table}
    \centering
    {\begin{tabular}{@{}r|cccc|ccc@{}} 
        \bottomrule
        &\multicolumn{4}{c|}{CUB-200-2023} & \multicolumn{3}{c}{HRSC-2023}  \\
        \hline
        Level label   & Order   & Family   & Species   & Total  & Type     & Sub-type & Total  \\
        \hline 
        Training set  & 1960    & 1054     & 2980      & 5994   & 434      & 838      & 1272  \\ 
        Test set      & 1895    & 1052     & 2847      & 5794   & 336      & 574      & 910  \\ 
        Total         & 3855    & 2106     & 5827      & 11788  & 770      & 1412     & 2182 \\ 
        \toprule
    \end{tabular}}
\end{table}

\subsection{Statistics of New Datasets}

The number of samples labeled to different levels in the hierarchy of CUB-200-2023 and HRSC-2023 datasets is presented in Table~\ref{statistic_table}. Since classifying objects at the leaf level (\textit{e.g.}, species) is more challenging than classifying objects at upper levels (\textit{e.g.}, order), we elaborately set up the distortions' strength so that samples labeled to leaf levels are more than samples labeled to superordinate levels.

\section{Method} 

\begin{figure}[h]
    \centering
    \includegraphics[width=0.99\linewidth]{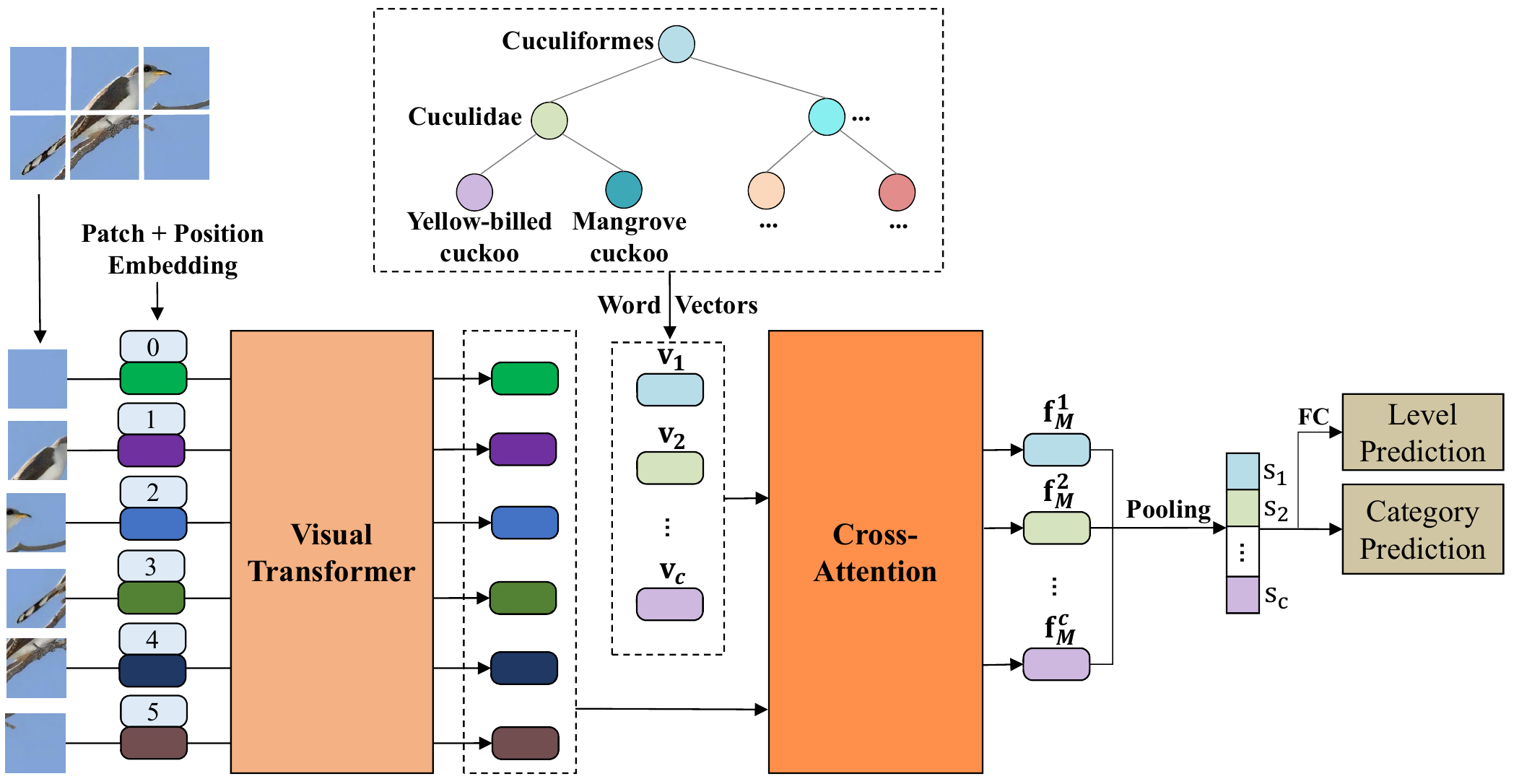} 
    \caption{The proposed Semantic Guided level-category Hybrid Prediction Network (SGHPN) consists of a visual transformer and a semantic guided cross-attention module.}
    \label{archi_fig}
\end{figure}

The overall framework of the proposed method is depicted in Fig.~\ref{archi_fig}, which mainly consists of two parts: the visual transformer (ViT) \cite{dosovitskiy2020image} and the semantic guided cross-attention module (SGCA). Given an image, ViT maps it to a sequence of token embeddings. Then, the categories in the label hierarchy are embedded into word vectors. In SGCA, the word vectors serve as queries, and the token embeddings from ViT serve as keys and values. For each category, SGCA utilizes the category semantics as guidance to learn the representation that focuses on the specific information of this category. The learned representations are then used for level prediction and category prediction.

\subsection{Vision Transformer as Feature Extractor}

Visual transformer (ViT)~\cite{dosovitskiy2020image} first converts an input image to a sequence of flattened patches. Then, these patches are mapped into a latent D-dimensional embedding space using a linear projection. To retain the position information, position embeddings are added to the patch embeddings. ViT contains $L$ blocks, with each block containing a multi-headed self-attention (MSA) and a multi-layer perceptron (MLP). The MLP contains two layers with a GELU non-linearity. Formally, the forward process of ViT can be represented as follows:
\begin{equation}
    \mathbf{Z}_0 = [\mathbf{E}^1_{patch}, \mathbf{E}^2_{patch}, \cdot \cdot \cdot, \mathbf{E}^N_{patch}] + \mathbf{E}_{pos}; 
\end{equation}
\begin{equation}
    \mathbf{Z}'_l = {\rm MSA}({\rm LN}(\mathbf{Z}'_{l-1})) + \mathbf{Z}'_{l-1}, \quad l = 1, ..., L;
\end{equation}
\begin{equation}
    \mathbf{Z}_l = {\rm MLP}({\rm LN}(\mathbf{Z}'_{l})) + \mathbf{Z}'_{l}, \quad l = 1, ..., L;
\end{equation}
where $\mathbf{E}^i_{patch} \in \mathbb{R}^{1 \times D}$ ($i=1, ..., N$) denotes the patch embeddings and $\mathbf{E}_{pos} \in \mathbb{R}^{N \times D}$ means position embedding. $N$ is the number of patches. ${\rm LN}(\cdot)$ stands for the layer normalization~\cite{ba2016layer}.

\subsection{Generating Word Embedding for Each Category}

For each label in the label hierarchy, we extract a D-dimensional semantic-embedding vector by leveraging on GloVe~\cite{pennington2014glove}. GloVe is an unsupervised learning algorithm for obtaining vector space representations of words. Its training is performed on aggregated global word-word co-occurrence statistics from corpora. The resulting representations capture the fine-grained semantics and show interesting linear substructures of the word vector space. GloVe's authors have released four text files\footnote{https://nlp.stanford.edu/projects/glove/.} with word vectors trained on different massive web datasets. In this work, we use the word vectors trained on ``Wikipedia 2014\footnote{https://dumps.wikimedia.org/enwiki/20140102/.} + Gigaword 5\footnote{https://catalog.ldc.upenn.edu/LDC2011T07.}'' dataset that contains about 6 billion tokens. It has 400k trained word vectors formatted as a dictionary, with items arranged as `\emph{word: word vector}'. These trained word vectors cover all the vocabulary of labels in the CUB-200-2023 and HRSC-2023 datasets. For each label, we perform a dictionary lookup to obtain its word vector. We denote these labels' embedding vectors as:
\begin{equation}
    \mathbf{V} = (\mathbf{v}_1, \mathbf{v}_2, ..., \mathbf{v}_c),
\end{equation}
where $c$ is the number of categories.

\begin{figure}[t]
    \centering
    \includegraphics[width=0.7\linewidth]{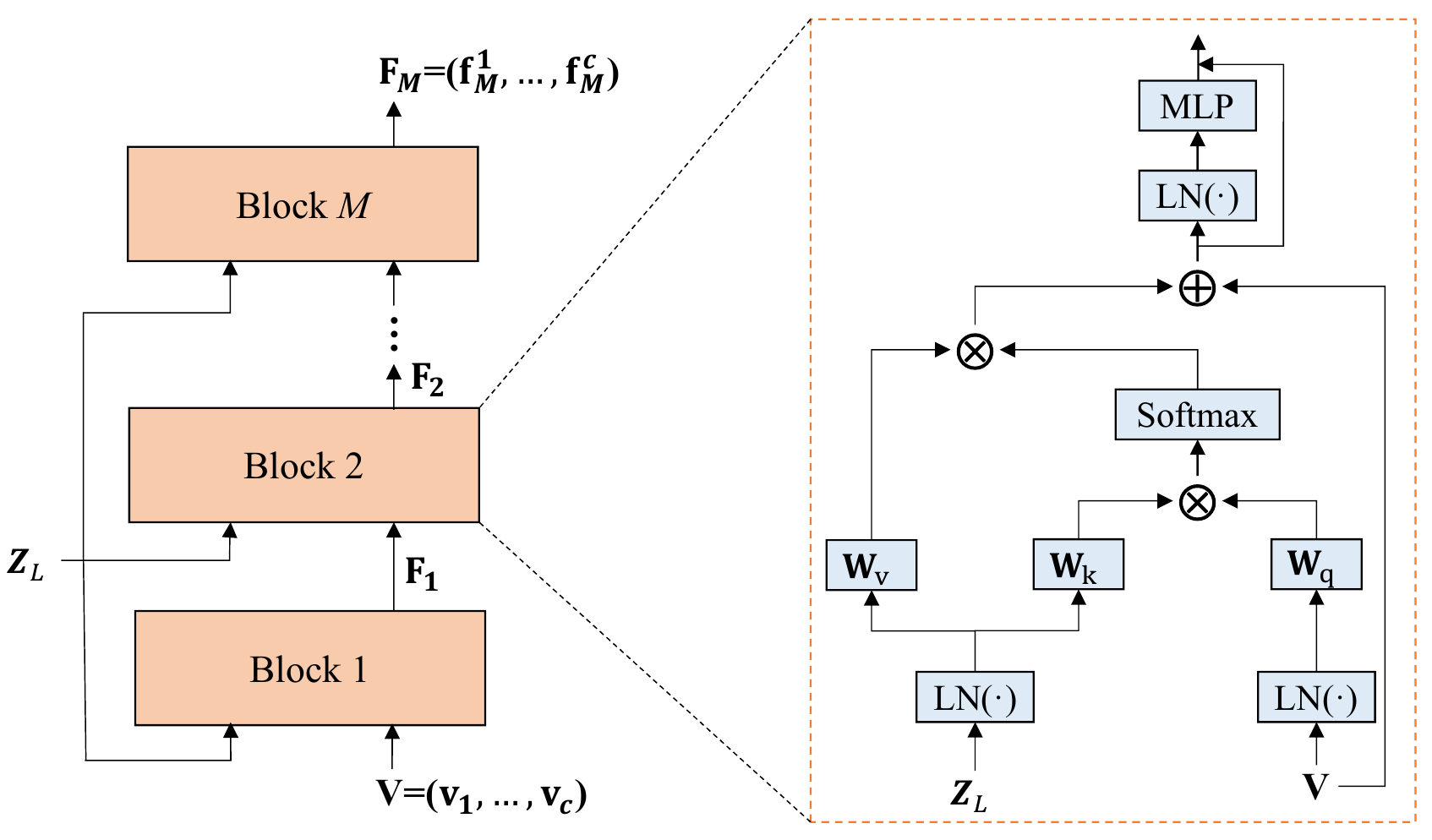} 
    \caption{The semantic guided cross-attention module (SGCA).}
    \label{decoder}
\end{figure}

\subsection{Semantic Guided Cross-Attention}

The Semantic Guided Cross-Attention module (SGCA) aims to learn representations under the guidance of category semantics to discover the correspondences between word embeddings and token embeddings via cross-attention learning. It is performed by using the word vectors as queries and the token vectors from ViT as keys. Here, we carry out the cross-attention via transformer decoder~\cite{vaswani2017attention}. SGCA comprises $M$ blocks, with each block containing a multi-head cross-attention (MCA) and an MLP. An illustration of the SGCA module is shown in Fig.~\ref{decoder}. Its forward process can be formulated as: 
\begin{equation}
\mathbf{F}'_l = {\rm MCA}({\rm LN}(\mathbf{F}_{l-1}, \mathbf{Z}_L)) + \mathbf{F}_{l-1}, \quad l = 1, ..., M;
\end{equation}
\begin{equation}
\mathbf{F}_l = {\rm MLP}({\rm LN}(\mathbf{F}'_{l}))+\mathbf{F}'_{l}, \quad l = 1, ..., M;
\end{equation}
where $\mathbf{F}_{0}=\mathbf{V}$ stands for the labels' word vectors, and $\mathbf{Z}_L$ is the token embeddings from ViT.

The SGCA module outputs $c$ token embeddings, \textit{i.e.,} $\mathbf{F}_M = (\mathbf{f}^1_M, \mathbf{f}^2_M, ..., \mathbf{f}^c_M)$. For each embedding, we adopt a fully-connected layer to pool it into a score that indicates the probability of a class. We perform this process for all output token embeddings and obtain a score vector $\mathbf{s}=(s_1,...,s_c)$, among which the max score implies the model's class prediction. Then, for the level prediction, we deploy a fully-connected layer upon score vector $\mathbf{s}$. We denote it as \textit{level predictor}.

\subsection{Loss Function}

To emphasize the discrimination among categories at each level, we impose the cross-entropy loss on each level. For convenience, we take the CUB-200-2023 dataset as an example and denote the cross-entropy loss on each level as $\mathcal{L}_{CE\_order}$, $\mathcal{L}_{CE\_family}$, $\mathcal{L}_{CE\_species}$, respectively. 
For the level prediction, as the predicted level only belongs to one of three mutually exclusive levels, we use cross-entropy to compute the level prediction error. The total loss can be defined as:

\begin{equation}
    \setlength\abovedisplayskip{0.1pt}   
    \setlength\belowdisplayskip{0.9pt}
    \mathcal{L} =
        \begin{cases}
            \mathcal{L}_{CE\_level} + \lambda \cdot \mathcal{L}_{CE\_order}, \qquad\qquad\ \ \ \text{if sample is in order nodes;}\\
            \mathcal{L}_{CE\_level} + \lambda \cdot (\mathcal{L}_{CE\_order} + \mathcal{L}_{CE\_family}), \\
            \qquad\qquad\qquad\qquad\quad\qquad\qquad\qquad\quad\ \text{if sample is in family nodes;}\\
            \mathcal{L}_{CE\_level} + \lambda \cdot (\mathcal{L}_{CE\_order} + \mathcal{L}_{CE\_family} + \mathcal{L}_{CE\_species}), \\
            \qquad\qquad\qquad\qquad\quad\qquad\qquad\qquad\quad\ \text{if sample is in species nodes;}
        \end{cases}
\end{equation}
where the parameter $\lambda$ balances the level prediction loss and the category prediction loss. We empirically set this parameter as $\lambda = 1$.

\section{Experiments} 

\subsection{Implementation Details}
 
We implement the model in PyTorch and optimize it using SGD with momentum of 0.9. For CUB-200-2023 dataset, the input images are resized to 448$\times$448, and the batch size is 16. The model is trained for 100 epochs with a learning rate of 0.0001. For HRSC-2023 dataset, the input images are resized to 224$\times$224, and the batch size is 32. The model is trained for 50 epochs with a learning rate of 0.002. Common training augmentation approaches are applied for both datasets, including horizontal flipping and random cropping (random cropping for training and center cropping for testing). The code and new datasets will be made publicly accessible.

\subsection{Evaluation Metrics}  \label{eval_metric}  
  
\begin{figure}[t]
    \centering
    \includegraphics[width=0.85\linewidth]{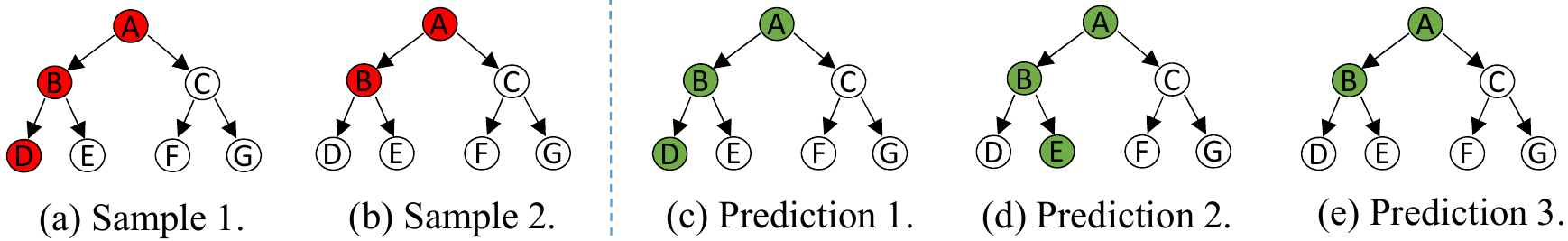} 
    \caption{The illustrative samples and predictions. The red nodes on the left are the true classes of two datapoints. The green nodes on the right are the predicted classes of three prediction cases.}
    \label{example_fig}
\end{figure}

For hierarchical classification, the hierarchical relationships should be considered when evaluating the performances of algorithms~\cite{Kosmopoulos2014EvaluationMF,brucker2011empirical}. The \textit{Symmetric Difference Loss ($SDL$)}, \textit{Hierarchical Precision ($P_H$)}, and \textit{Hierarchical Recall ($R_H$)}~\cite{Kosmopoulos2014EvaluationMF} are three typically used hierarchical metrics, which are adopted in this work for model evaluation. Their definition are described as below, and they are visually illustrated by the toy examples in Fig.~\ref{example_fig} and Table~\ref{all_score_table}.
  
$SDL$ is calculated as follows:
\begin{equation}
    SDL = |(S \backslash \hat{S}) \cup (\hat{S} \backslash S)|,
\end{equation}
where $|\cdot|$ is the cardinality of a set. $S$ and $\hat{S}$ are the ground truth set and predicted set, respectively. For example, in Fig.~\ref{example_fig}(a), $S$ is (A, B, D), and in Fig.~\ref{example_fig}(d), $\hat{S}$ is (A, B, E). $SDL$ measures the degree of errors with the difference between the ground truth set and predicted set. Lower $SDL$ value means better performance.

$P_H$ is defined as follows:
\begin{equation}
    P_H = \frac{|S \cap \hat{S}|}{|\hat{S}|}.
\end{equation}
$P_H$ measures how precise the algorithm is in classification. Since the denominator is $|\hat{S}|$, $P_H$ punishes the over-specific predictions. Higher $P_H$ value indicates better performance.

$R_H$ is defined as follows:
\begin{equation}
    R_H = \frac{|S \cap \hat{S}|}{|S|}.
\end{equation}
$R_H$ measures how sensitive the algorithm is to the positive case. Since the denominator is $|S|$, which is fixed for a datapoint, and the numerator is $|S \cap \hat{S}|$, $R_H$ penalizes the under-specific predictions. Higher $R_H$ value means better performance.

\begin{table}[t] \footnotesize
    \caption{$SDL$, $P_H$, and $R_H$ of three predictions over two samples from Fig.~\ref{example_fig}.}
    \label{all_score_table}
    \centering
    {\begin{tabular}{@{}cccc|cccc@{}} 
        \bottomrule
        & $SDL$  & $P_H$   & $R_H$     &              & $SDL$    & $P_H$   & $R_H$ \\
        \hline
        P1 on S1*        & 0      & 1.00    & 1.00      & P1 on S2     & 1     & 0.67    & 1.00 \\
        P2 on S1         & 2      & 0.67    & 0.67      & P2 on S2     & 1     & 0.67    & 1.00 \\
        P3 on S1         & 1      & 1.00    & 0.67      & P3 on S2     & 0     & 1.00    & 1.00 \\
        \toprule
        \multicolumn{8}{l}{* ``P1 on S1" means ``Prediction 1 on Sample 1".}
    \end{tabular}}
\end{table}

\subsection{Ablation Study} 

\textbf{Contribution of SGCA.} We evaluate the contribution of SGCA by comparing the model performance with and without this module. Without this module, the model is a plain ViT, and a [cls] token is prepended to the patch tokens for classification, as done in previous works\cite{dosovitskiy2020image, devlin2018bert}. Table~\ref{ablation} compares the model performance on CUB-200-2023 dataset with and without SGCA module. It shows that ``With SGCA" performs better than ``Without SGCA", which validates the effectiveness of SGCA and supports our motivation that we use the semantics as attention guidance to extract the category semantic-specific representation.

\begin{table}[t] \footnotesize
    \caption{$SDL$, $P_H$, and $R_H$ results on CUB-200-2023 dataset with and without SGCA module.}
    \label{ablation}
    \centering
    {\begin{tabular}{@{}c|cccc@{}} 
        \bottomrule
        & $SDL\downarrow$  & $P_H$(\%)$\uparrow$  & $R_H$(\%)$\uparrow$  \\
        \hline
        Without SGCA        & 0.5641   & 90.53   & 91.78   \\
        With SGCA           & \textbf{0.5240}   & \textbf{90.71}   & \textbf{92.84}   \\
        \toprule
    \end{tabular}}
\end{table}

\begin{figure}[t]
    \centering
    \subfigure[The predicted level is species.]{\includegraphics[width=1.75in]{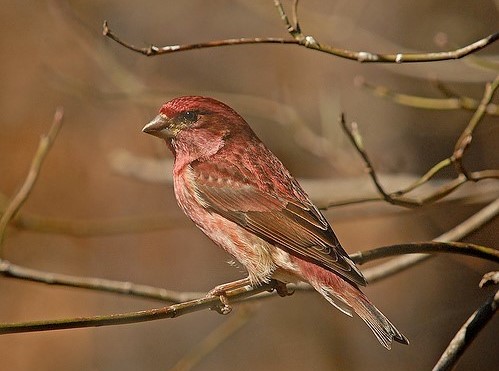}}
    \subfigure[The predicted level is species.]{\includegraphics[width=1.75in]{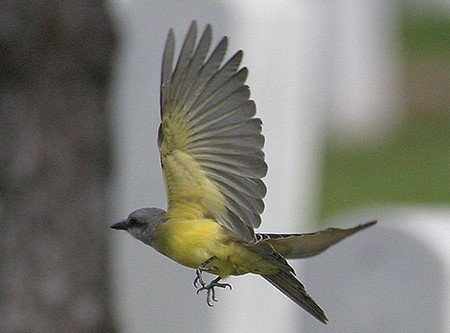}}\\
    \subfigure[The predicted level is family.]{\includegraphics[width=1.75in]{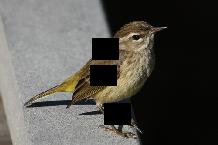}}
    \subfigure[The predicted level is family.]{\includegraphics[width=1.75in]{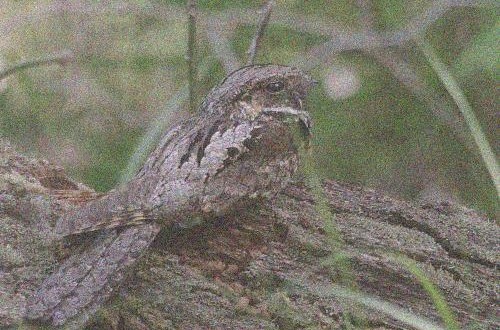}}\\
    \subfigure[The predicted level is order.]{\includegraphics[width=1.75in]{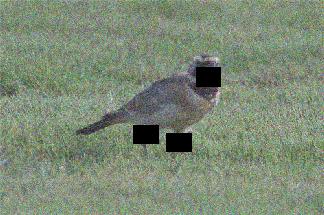}}
    \subfigure[The predicted level is order.]{\includegraphics[width=1.75in]{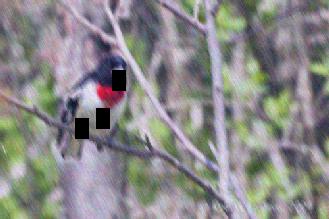}}\\
    \caption{Predictions of our level predictor on six illustrative examples from the test set of CUB-200-2023 dataset. From the first row to the third row, the images' quality is from high to low. Our level predictor is capable of selecting fine-grained level for high-quality images and coarse-grained level for low-quality images.} 
    \label{levels_fig}
\end{figure}

\noindent\textbf{Effectiveness of Level Predictor}. A good level predictor should be able to select an appropriate level for the test image according to its image quality. Fig.~\ref{levels_fig} shows the prediction results of our proposed level predictor over some images from the test set of the CUB-200-2023 dataset. We can observe that Fig.~\ref{levels_fig} (a) and (b) are in relatively good condition, our level predictor selects a leaf level (\textit{i.e.}, species) for them. If images are of poor quality, our level predictor accordingly gives up striving for the lower level. Instead, it selects the upper level. For instance, it predicts the uppermost level (\textit{i.e.}, order) for Fig.~\ref{levels_fig} (e) and (f) because they are severely impaired.

\begin{table}[t] \footnotesize
    \caption{Comparison with state-of-the-art methods on CUB-200-2023 dataset and HRSC-2023 dataset.}
    \label{comparison_table}
    \centering
    {\begin{tabular}{@{}c|ccc|ccc@{}} 
        \bottomrule
        \multirow{2}{*}{Methods} & \multicolumn{3}{c|}{CUB-200-2023}  & \multicolumn{3}{c}{HRSC-2023} \\
        \cline{2-7}
         & $SDL\downarrow$  & $P_H$(\%)$\uparrow$  & $R_H$(\%)$\uparrow$  & $SDL\downarrow$  & $P_H$(\%)$\uparrow$  & $R_H$(\%)$\uparrow$ \\
        \hline
        HMCN~\cite{wehrmann2018hierarchical}   & 1.7235   & 68.24   & 68.49   & 0.4253   & 88.57   & 89.62  \\
        HMC-LMLP~\cite{cerri2016reduction}     & 1.4976   & 73.09   & 71.57   & 0.4582   & 86.87   & 88.79 \\
        Chang et al. \cite{chang2021your}       & 0.8284   & 83.68   & 81.95   & 0.3868   & 90.49   & 89.12  \\
        Chen et al. \cite{chen2022label}       & 0.8107   & 85.42   & 84.41   & 0.3758   & 90.88   & 89.45  \\
        SGHPN(\textbf{Ours})  & \textbf{0.5240}   & \textbf{90.71}   & \textbf{92.84}   & \textbf{0.3516}  & \textbf{91.52}   & \textbf{93.42}   \\
        \toprule
    \end{tabular}}
\end{table}

\subsection{Comparison with State-of-the-art Methods}  

In this section, we compare our proposed network to four state-of-the-art HC methods: HMCN\cite{wehrmann2018hierarchical}, HMC-LMLP\cite{cerri2016reduction}, and the approaches proposed by Chang et al.\cite{chang2021your} and Chen et al.\cite{chen2022label}. 
These methods can only predict category, while our focus is on level-category hybrid prediction. Hence, we adapt them into level-category hybrid prediction models by using them as backbone networks respectively.
HMCN~\cite{wehrmann2018hierarchical} simultaneously optimizes local and global loss functions for discovering local hierarchical class relationships and global information from the entire class hierarchy. We add a classifier head that uses both local and  global information for level prediction. 
HMC-LMLP~\cite{cerri2016reduction} proposes to train a chain of hidden FC layers where each FC layer corresponds to a level. The output of the previously trained FC layer is fed into the next FC layer to enhance the representation of the instance. 
For level prediction, we add a separate classifier that combines output features from all levels. 
The model proposed by Chang et al.~\cite{chang2021your} leverages level-specific classification heads to disentangle fine-grained features from coarse-level ones and allows finer-grained features to participate in coarser-grained label predictions. We add a classifier head that uses both fine-grained and coarse-grained features to output the level prediction. 
The model of Chen et al.\cite{chen2022label} is equipped with a hierarchical feature interaction module that contains granularity-specific blocks to extract the specialized feature for each hierarchical level. We deploy a classier head that uses features from all blocks for the level prediction.

Table~\ref{comparison_table} presents the comparisons between our method and the state-of-the-art HC approaches. Our method achieves the best performance on CUB-200-2023 and HRSC-2023 datasets.


\section{Conclusion} 


Images that are impaired by noise, occlusion, blur, or low resolution may not provide sufficient information for lower-level classification. Thus, we assume that each image should belong to an appropriate level in the class hierarchy according to its individual image quality. Accordingly, we propose a novel Semantic Guided level-category Hybrid Prediction Network (SGHPN) that can predict the level and the category jointly in an end-to-end manner. 
To evaluate the proposed method, we construct datasets in which images are at a broad range of quality and thus are labeled to different levels in the class hierarchy. Experimental results demonstrate the effectiveness of our proposed method by comparing it to the state-of-the-art HC approaches.

\section*{Acknowledgments}
This work was supported by the National Natural Science Foundation of China under Grant 62071421.

\end{CJK}

\bibliographystyle{ws-ijwmip}
\bibliography{refs}

\end{document}